\documentclass{article}
\usepackage[preprint]{neurips_2023}
\usepackage[utf8]{inputenc} 
\usepackage[T1]{fontenc}    
\usepackage{hyperref}       
\usepackage{url}            
\usepackage{booktabs}       
\usepackage{amsfonts}       
\usepackage{nicefrac}       
\usepackage{microtype}      
\usepackage{xcolor}         
\usepackage{microtype}
\usepackage{graphicx}
\usepackage{subfigure}
\usepackage{booktabs} 
\usepackage{multirow}
\usepackage{hyperref}
\usepackage{amsmath}
\usepackage{amssymb}
\usepackage{mathtools}
\usepackage{amsthm}
\usepackage{adjustbox}
\usepackage{graphicx}
\usepackage{array}
\usepackage{subfigure}
\usepackage{verbatim}
\usepackage{bm}     
\usepackage{enumitem}       
\usepackage{makecell}
\usepackage[capitalize,noabbrev]{cleveref}
\usepackage{ragged2e}

\title{Hunyuan-Large: An Open-Source MoE Model with 52 Billion Activated Parameters by Tencent}

\author{
}

\author{\hyperref[sec:contributors]{Tencent Hunyuan Team}}
\begin{document}
\maketitle
\begin{abstract}
In this paper, we introduce Hunyuan-Large, which is currently the largest open-source Transformer-based mixture of experts model, with a total of 389 billion parameters and 52 billion activation parameters, capable of handling up to 256K tokens. 
We conduct a thorough evaluation of Hunyuan-Large's superior performance across various benchmarks including language understanding and generation, logical reasoning, mathematical problem-solving, coding, long-context, and aggregated tasks, where it outperforms LLama3.1-70B and exhibits comparable performance when compared to the significantly larger LLama3.1-405B model. 
Key practice of Hunyuan-Large include large-scale synthetic data that is orders larger than in previous literature, a mixed expert routing strategy, a key-value cache compression technique, and an expert-specific learning rate strategy. 
Additionally, we also investigate the scaling laws and learning rate schedule of mixture of experts models, providing valuable insights and guidances for future model development and optimization. 
The code and checkpoints of Hunyuan-Large are released to facilitate future innovations and applications. 

\textbf{Code}: \href{https://github.com/Tencent/Hunyuan-Large}{https://github.com/Tencent/Tencent-Hunyuan-Large} \\
\textbf{Models}: \href{https://huggingface.co/tencent/Tencent-Hunyuan-Large}{https://huggingface.co/tencent/Tencent-Hunyuan-Large}
\end{abstract}

\section{Introduction}
In recent years, Large language models (LLMs) have significantly advanced the field of artificial intelligence, proving their effectiveness across numerous fields such as NLP, CV, Speech, and AI4Science. Starting from the emergence of ChatGPT \citep{OpenAI2022ChatGPT}, lots of powerful LLMs have bloomed \citep{achiam2023gpt,Gemini2023gemini,touvron2023llama,OpenAI2024GPT4o,dubey2024llama,QWen2024qwen2-5}, which inexorably bring in new ways for people to collect and process information, broadly impacting our daily lives. As the demand for more sophisticated AI systems continues to grow, researchers are exploring new techniques and paradigms to push the boundaries of model size and performance. One approach that stands out is the Mixture of Experts (MoE) model, which synergizes multiple specialized submodels to deliver superior performance in diverse tasks with dynamic activated experts \citep{lepikhin2020gshard,fedus2022switch,wang2024hmoe}, 
achieving more efficient training and inference. There is a current trend observed that more and more MoE-structured LLMs have been constructed and open-sourced to facilitate the LLM community \citep{mistral2024mixtral8x22,deepseek2024v2,yang2024qwen2,team2024jamba}.

Tencent's AI chatbot, Yuanbao (yuanbao.tencent.com), has also adopted MoE as the neural architecture of the trillion-parameter flagship LLM since February 2024. Due to its exceptional capabilities in reading, writing, and searching, the MoE-based Hunyuan model and Yuanbao chatbot are assisting users in working effortlessly and enjoying a more vibrant life. The MoE-powered Hunyuan models have also enhanced thousands of scenarios within Tencent's applications, enabling Tencent to better serve its billions of users. 

In addition to serving users with the premium models, another way that contributes to the community is open-sourcing. Open-source models can greatly promote the spreading of technology and flourishing development of applications, as exemplified by LLama, Mistral, Qwen, and Deepseek, among others. However, most open-source models are based on dense architectures, with only a very few models based on the MoE architecture with relatively small scale of parameters. In this work, we introduce Hunyuan-Large, a large Transformer-based MoE model, featuring an unprecedented 389 billion total parameters and 52 billion activated parameters, capable of handling up to 256K tokens. This model adopts the classical Transformer architecture \citep{vaswani2017attention} with MoE, containing a pre-training stage for acquiring fundamental capabilities and a post-training stage for task-specific instruction following, capability enhancement, and human preference alignment.
Hunyuan-Large supports conventional NLP abilities such as question answering, reasoning, reading comprehension, and specific LLM capabilities such as mathematics, coding, multi-turn interaction, and multilinguality.

We delve into the key technical innovations that have contributed to Hunyuan-Large's exceptional performance as follows.
\begin{itemize}[leftmargin=10pt]
\item \textbf{High-Quality Synthetic Data}. The broad usage of synthetic data improves the quality and diversity of training data, which enables the model to learn richer representations effectively and generalize better to unseen data. In total, Hunyuan-Large is pre-trained on 7T tokens, which contains nearly 1.5T tokens of high-quality and diverse synthetic data.
\item \textbf{Enhanced Model Structure}. We propose key-value (KV) cache compression, recycle routing, and expert-specific learning rate scaling strategies to enhance Hunyuan-Large. The reduction of KV cache overhead allows for more seamless deployment and scaling. Moreover, we adopt different learning rates for different shared/specialized experts with our recycle routing strategy, ensuring that each token can be utilized effectively during training and contributing to the overall performance. 
\item \textbf{Explorations on MoE Scaling Laws.} Additionally, we explore the scaling laws of MoE models as our guidelines, highlighting the relationship between model size, training data, and performance. This analysis offers insights into the foundational elements that contribute to the strong performance of Hunyuan-Large, but also provides valuable insights for future development and optimization of more powerful and larger MoE-structured LLMs.
\end{itemize}

To demonstrate the power of Hunyuan-Large, we conduct extensive experiments on diverse types of benchmarks in both English and Chinese, compared with the best-performing dense and MoE models having similar parameter sizes. We find that Hunyuan-Large is capable of handling various tasks including commonsense understanding, question answering, mathematics reasoning, coding, and aggregated tasks, achieving the overall best performance among existing open-source similar-scale LLMs. 
The pre-trained and post-trained Hunyuan-Large models are publicly released to facilitate the LLM community.

In the rest of this technical report, we will first give a detailed introduction to the pre-training stage of Hunyuan-Large, including its data and tokenizer, model structure, and pre-training recipes in Section \ref{sec:pre-training}. Next, we will describe our post-training in Section \ref{sec:post-training}, with details of our SFT and RLHF techniques. The comprehensive experimental results and in-depth analyses of Hunyuan-Large's pre-trained and post-trained models will be given in Section \ref{sec:evaluation}. Finally, the conclusion and future direction will be stated in Section \ref{sec:conclusion}.

\section{Pre-Training}
\label{sec:pre-training}

In this section, we will describe the details of pre-training Hunyuan-Large, including (a) data and tokenizer, where high-quality data largely contributes to the model performance, (b) model structure, consisting of our proposed KV cache compression, expert routing, and expert-specific learning rate scaling strategies, and (c) pre-training recipes, introducing the detailed pre-training schedule as well as our guidebook of explorations on MoE scaling laws. These techniques build the foundation of Hunyuan-Large's remarkable capability in pre-training.
\begin{figure}[!ht]
	\centering
 \includegraphics[page=1,width=0.98\columnwidth]{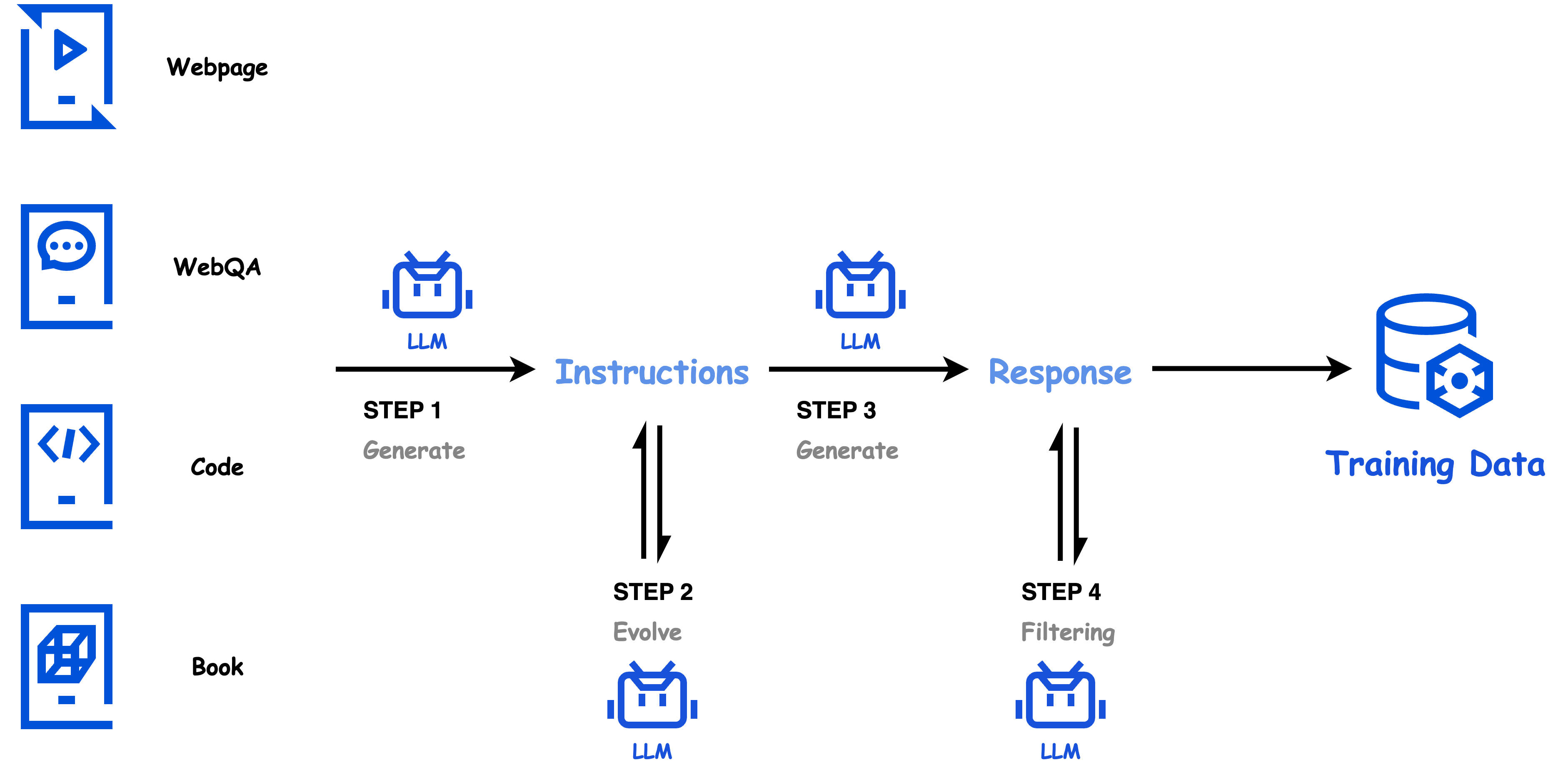}
	\caption{The four-step process of data synthesis in Hunyuan-Large's pre-training: (1) Instruction generation, (2) Instruction evolution, (3) Response generation, and (4) Response filtering.}
	\label{fig:data_syn}
\end{figure}
\subsection{Data and Tokenizer}

We first give the overview of our data, which is viewed as the fuel of our powerful model, with its preprocessing steps and data synthesis strategies essential for the quantity and quality of data. We also introduce the tokenizer employed for converting text data into an appropriate format suitable for Hunyuan-Large.

\subsubsection{Data Processing and Synthesis}
\label{sec:data}

To start with, we provide a brief overview of the used pre-training data, and then delve deeper into the specifics of our synthetic data generation process, which is essential for acquiring capabilities also verified in various LLMs \citep{dubey2024llama,abdin2024phi3,liu2024best}.

\textbf{Data Overview and Processing.}
We aim to create a high-quality, safe, and diverse training dataset for pre-training, primarily consisting of Chinese and English languages for practical demands. We filter the data based on criteria such as writing quality, educational value, and toxicity to ensure its high quality. Additionally, we anonymize all privacy-sensitive data and other harmful data. We have also implemented an elaborate system of category labels, which allows us to flexibly adjust the proportions of various types of data in the training dataset. 

\textbf{Data Synthesis.}
Besides the existing natural text corpus, we construct large amounts of synthetic data to specifically boost the knowledge acquisition against the relative capability deficiency merely learned from natural data. To make full use of synthetic data to enhance model performance, we mainly focus on the mathematics, coding, low-resource, and high-educational-value fields that are good supplements to naturally distributed corpus, meeting three key requirements of quality, diversity, and quantity.

As shown in Figure \ref{fig:data_syn}, we synthesize high-quality instruction data through a four-step process, including instruction generation, instruction evolution, response generation, and response filtering.
\begin{itemize}[leftmargin=10pt]
    \item Step 1: \textbf{Instruction Generation}. To ensure the diversity of instructions, we use high-quality, knowledge-rich data sources such as web pages, web-based question-answering data, code repositories, books, and other resources as seeds. Cooperating with diverse instruction generation prompts, these seeds enable us to generate a wide variety of instructions that cover various domains with different desired instruction styles and complexities.
    \item Step 2: \textbf{Instruction Evolution}. To further improve the quality of these initial instructions, we refine them by three guidelines: (a) Enhancing their clarity and informativeness. (b) Expanding low-resource domain instructions through self-instruct augmentation. (c) Evolving the instructions to increase their difficulty levels. These evolved high-quality and challenging instructions enable our model to benefit more efficiently from synthetic data to cross the original capability boundaries.
    \item Step 3: \textbf{Response Generation}. We utilize several specialized models to generate informative and accurate answers for the above evolved instructions. These models vary in size and are well-designed specialized models to synthesize expert-level responses for instructions in various domains.
    \item Step 4: \textbf{Response Filtering}. To filter the synthetic instruction-response pairs, we employ a critique model and conduct self-consistency checks, in which we generate multiple answers to perform self-consistency filtering for tasks such as objective question-answering tasks, ensuring the reliability and accuracy. This process allows us to effectively remove any low-quality or inconsistent data, ensuring the utilization of high-quality text in pre-training.

\end{itemize}

\subsubsection{Tokenizer}
The tokenizer is a vital component for effectiveness and efficiency in pre-training, which should balance two critical factors: (a) achieving a high compression rate for efficient training and inference, and (b) maintaining an appropriately large vocabulary to ensure adequate learning of each word embedding. 
In Hunyuan-Large, we carefully consider both aspects and employ a vocabulary consisting of 128K tokens. This token vocabulary is a combination of 100K tokens from the tiktoken tokenizer \citep{OpenAI2022Tiktoken} and an additional 28K tokens specifically designed to enhance Chinese language support. Notably, when compared to the LLama3.1 tokenizer, our new tokenizer exhibits improved compression rates, increasing from 2.78 to 3.13 characters per token.

\subsection{Model Structure}
Hunyuan-Large is equipped with superior model structure and training strategies to achieve impressive LLM capabilities. We first show the overview of model architecture and hyper-parameters, and then delve into the KV cache compression, expert routing strategy, and expert-specific learning rate scaling used in our model with details.

\subsubsection{Overview of Hunyuan-Large}
The model structure of Hunyuan-Large mainly follows the classical MoE structure that uses multiple experts to replace the original FFN in Transformer. Tokens will be assigned to different experts, and only a small ratio of experts will be activated in training. Hunyuan-Large consists of both shared and specialized experts. We use Rotary Position Embedding (RoPE) for position learning \citep{su2024roformer} and SwiGLU for activation \citep{shazeer2020glu}.
Table \ref{tab:overall_para} displays the overview of our model's architecture and key hyper-parameters.

\begin{table}[!hbtp]
    \centering
    \caption{Overview of the architecture and key hyper-parameters of Hunyuan-Large. This model has 389B total parameters and 52B activated parameters. There are 1 shared expert and 1 specialized expert activated for each token.}
        \begin{tabular}{l|c}
            \toprule
            \textbf{Configuration} & \textbf{Hunyuan-Large} \\
            \midrule
            \# Layers & 64\\
            \# Attention Heads & 80\\
            \# Key/Value Heads & 8\\
            \# Shared Experts & 1\\
            \# Specialized Experts & 16\\
            \# Activated Specialized Experts & 1\\
            \# Trained Tokens & 7T\\
            Activation Function & SwiGLU\\
            Vocabulary Size & 128K\\
            Hidden Size & 6,400\\
            \bottomrule
        \end{tabular}

    \label{tab:overall_para}
\end{table}

\subsubsection{KV Cache Compression}
To alleviate memory pressure of KV cache and reduce the cost during inference, we jointly integrate two classical strategies for KV cache compression: (a) Grouped-Query Attention (GQA) \citep{ainslie2023gqa}, which uses an intermediate number of KV heads to form head groups, compressing KV cache from the head aspect, and (b) Cross-Layer Attention (CLA) \citep{brandon2024reducing}, which shares the KV cache between adjacent layers, compressing KV cache from the layer aspect. In Hunyuan-Large, we set $8$ groups of KV heads for GQA, and share KV cache every $2$ layers, jointly considering both effectiveness and efficiency. Table \ref{tab:kv_cache_comparison} presents a comparison of the KV cache memory usage across different mechanisms. The adopted GQA+CLA technique in Hunyuan-Large saves nearly 95\% KV cache in total compared to the original MHA mechanism, significantly improving the inference efficiency without much side effect on model performance.
\begin{table}[!hbtp]
    \centering
    \caption{Comparisons of KV cache memory (in bytes on bf16) for different attention mechanisms. The attention mechanisms include Multi-Head Attention (MHA), Grouped-Query Attention (GQA), Multi-Query Attention (MQA), Cross-Layer Attention (CLA), and GQA+CLA (the final setting in Hunyuan-Large). \(n_h\), \(d_h\), \(l\), and \(n_g\) represent the number of attention heads, the dimension per head, the number of layers, and the number of groups in GQA (\(n_g\)<\(n_h\)), respectively. Our CLA shares the KV cache every 2 layers.}
        \begin{tabular}{l|c} 
            \toprule
            \textbf{Attention Mechanism} & \textbf{KV Cache Memory} \\
            \midrule
            MHA & $4n_h d_h l$ \\
            GQA & $4n_g d_h l$ \\
            MQA & $4d_h l$ \\
            CLA & $2n_h d_h l$ \\
            GQA+CLA & $2n_g d_h l$ \\
            \bottomrule
        \end{tabular}
    \label{tab:kv_cache_comparison}
\end{table}

\subsubsection{Expert Routing Strategy}
\begin{figure}[hbtp]
    \centering
    \subfigure[Traditional Top-k Routing.]{
    \includegraphics[page=2,width=.49\textwidth]{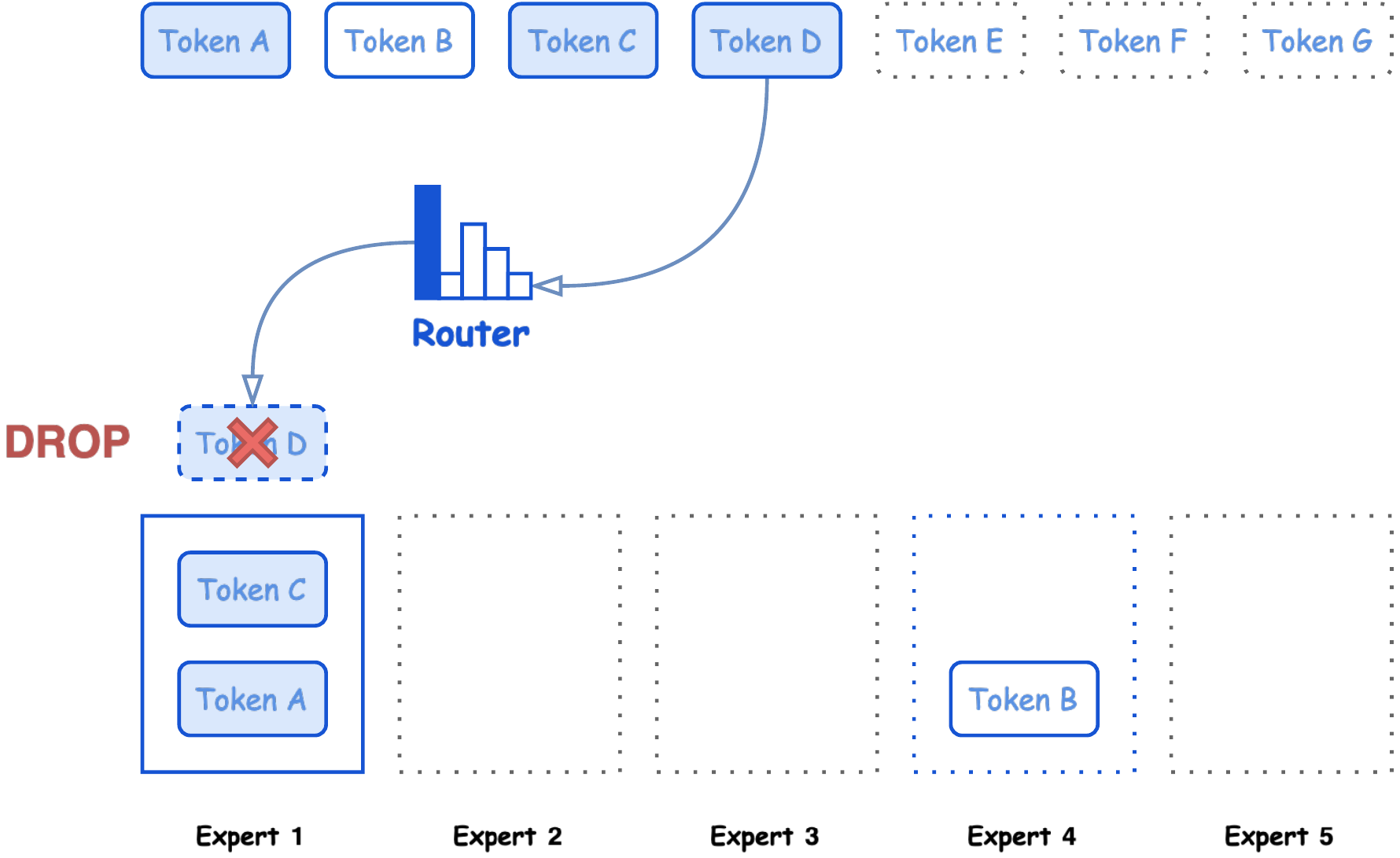}
    \label{fig:recycle_routing_a}
    }
    \subfigure[Recycle Routing.]{
    \includegraphics[page=3,width=.44\textwidth]{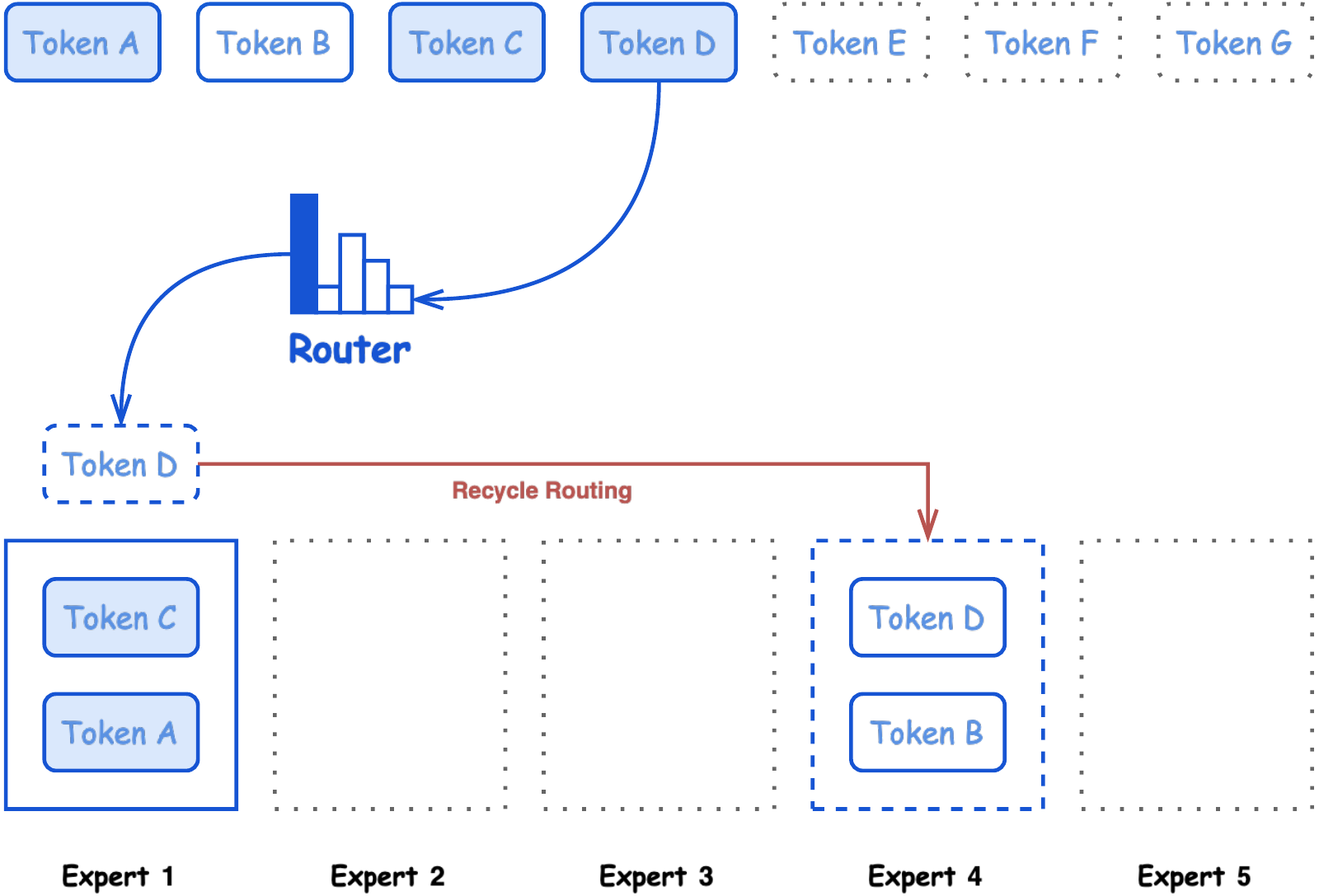}
    \label{fig:recycle_routing_b}
    }
    \caption{An illustration of the recycle routing strategy in Hunyuan-Large, where each expert's maximum capacity is set to 2. Token D, which was initially allocated to the overloaded Expert 1, is reassigned to a randomly selected Expert 4. This approach helps alleviate the potential loss of valuable information. In traditional routing strategies, tokens from overloaded experts would be dropped as shown in (a). However, our strategy involves randomly reassigning these tokens to other experts, as demonstrated in (b), where Token D is routed to Expert 4.}
    \label{fig:recycle_routing}
\end{figure}

\noindent
\textbf{Shared and Specialized Experts.}
The expert routing strategy of MoE is essential to efficiently activate each expert's capability while maintaining a relatively balanced load. Conventional routing strategies, such as the classical top-k routing strategy, selects the top-k scoring experts to process each token \citep{jiang2024mixtral,qwen_moe}. Hunyuan-Large adopts a mixed routing strategy, that uses both a shared expert consumed by all tokens and several routable experts employing the classical top-k routing strategy~\footnote{This mixed routing strategy was first introduced in our closed-source trillion-parameter model (training starts from November, 2023) concurrently to Deepseek v2.}. 
Hunyuan-Large sets 1 expert as the shared expert to capture the common knowledge required by all tokens. Besides, 16 specialized experts are allocated to dynamically learn domain-specific knowledge, activating the top-$1$ scoring specialized expert for each token.
\noindent
\textbf{Recycle Routing.}
Conventional top-k routing often cooperates with a capacity factor that defines the maximum load of an expert in MoE, where tokens of overloaded experts are discarded during training. A larger capacity factor results in less dropped tokens but reduced training efficiency.
Excessive token dropping may cause the loss of crucial information, which in turn negatively impacts training stability. To address this problem and achieve more balanced training in efficiency and stability, we develop a new recycle routing strategy for tokens discarded during the original top-k routing process, as displayed in Figure \ref{fig:recycle_routing}. This technique entails an additional random allocation for tokens originally routed to overloaded experts to other specialized experts which have not exceeded their capacity. This approach strives to preserve vital information while simultaneously optimizing training efficiency, thus ensuring the overall effectiveness and efficiency of model training.

\subsubsection{Expert-Specific Learning Rate Scaling}
We adopt AdamW \citep{loshchilov2019decoupled} as our optimizer. To expedite training, we can increase the learning rate in tandem with the growth of batch size in pre-training. 
Previous work has explored the square root scaling \citep{krizhevsky2014one} or linear scaling \citep{goyal2017accurate} when discovering the optimal learning rate based on the batch size for SGD-style optimizers. Recent work has elucidated a more appropriate connection between the optimal learning rates and batch sizes for Adam-style optimizers in LLMs. According to \citet{li2024surge}, the optimal learning rate $\epsilon_{opt}(B)$ for a batch size $B$ is calculated as:
\begin{equation}
\begin{split}
\epsilon_{opt}(B) = \frac{2\epsilon_{max}}{\sqrt{\frac{\mathcal{B}_{noise}}{B}} + \sqrt{\frac{B}{\mathcal{B}_{noise}}}}.
\end{split}
\end{equation}
Here, $\epsilon_{max}$ represents the learning rate of AdamW. $\mathcal{B}_{noise}$ indicates the trade-off point between training speed and data efficiency noted in \citet{kaplan2020scaling}.

However, in Hunyuan-Large, different experts are imbalanced in the aspect of trained tokens (e.g., comparing the shared expert with other specialized experts). The number of tokens processed by each expert during a single iteration will vary, which indicates that each expert will experience a different effective batch size within one training iteration. Hence, it is essential to necessitate expert-specific learning rates to optimize training efficiency. Considering the load balance losses, we could safely assume that different specialized experts have approximately similar numbers of effectively trained tokens.
Specifically for specialized experts, the effective batch size should be roughly divided by the number of specialized experts, resulting in their optimal learning rate being expressed as $\epsilon_{opt}({B}/{n})$ (we activate 1 in 16 specialized experts and thus $n=16$).
The learning rate scaling ratio between the shared and specialized experts is $\epsilon_{opt}(B) / \epsilon_{opt}({B}/{n})$, which is approximately $0.31$ in our setting. Consequently, when configuring the learning rate for Hunyuan-Large, we assign the optimal $\epsilon_{opt}(B)$ for the shared expert, and deliberately scale down the learning rate of specialized experts in accordance with this ratio $\epsilon_{opt}(B) / \epsilon_{opt}({B}/{n})$.

\subsection{Pre-Training Recipes}
The effectiveness of LLM pre-training is not solely determined by the dataset and model structure, but also significantly ascribed to the pre-training recipes obtained from empirical experiments. We first explore the scaling laws of MoE functioned as a guidebook for our model design. Secondly, we introduce the detailed process of annealing and long-context pre-training, which further enhance LLM's capability.

\subsubsection{MoE Scaling Law}
\begin{figure}[!hbtp]
	\centering
	\includegraphics[width=0.88\columnwidth]{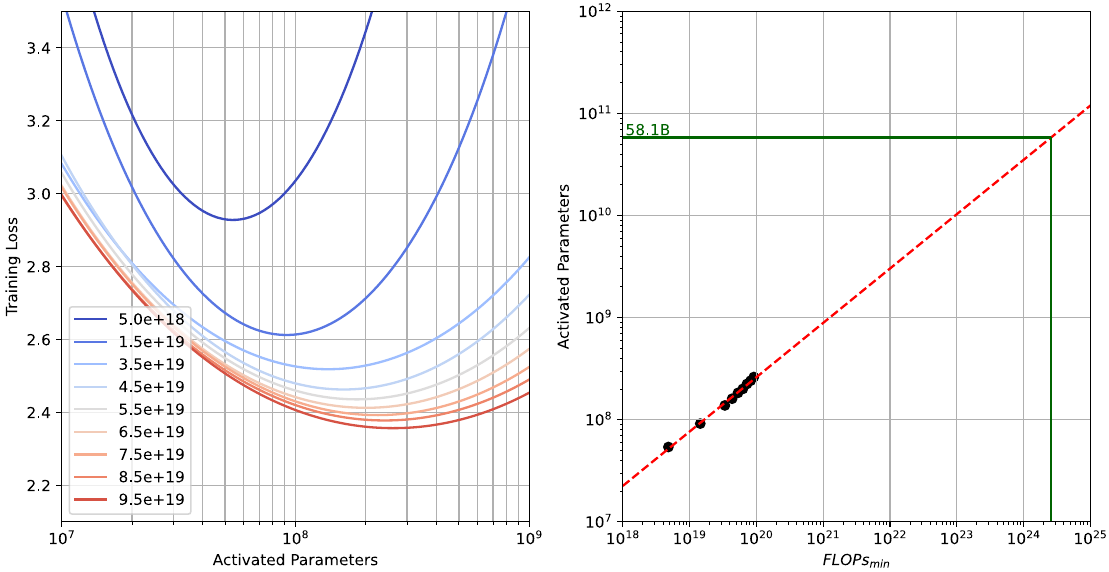}
	\caption{Using quadratic polynomial fitting, we obtain the scaling law of the optimal number of activation parameters under different minimum compute budgets.}
	\label{fig:IsoFLOP_curve-AP}
\end{figure}

Initially, we investigate the scaling laws of MoE models to identify optimal settings and gain insights before pre-training. Typically, the training compute budget for dense models is estimated using $C = 6ND$, where $N$ represents the number of parameters and $D$ denotes the training tokens. However, for MoE models with longer sequences (e.g., 8K, 32K, and 256K), the compute budget formula varies due to attention complexity and sparse activation. Upon meticulous computation, we ascertain the precise compute budget $C$ for MoE models, where $N$ in our formula represents the number of activated parameters, is as follows:
\begin{equation}
\begin{split}
C \approx 9.59 ND + 2.3 \times 10^8 D.
\end{split}
\end{equation}

Drawing on the insights of \citet{kaplan2020scaling} and \citet{li2024surge}, we acknowledge that batch size $B$ has a significant impact on compute budget $C$ during training. To isolate this effect and derive precise estimates, we employ the critical batch size $B_{crit}(L)$, which optimizes the trade-off between time and computational efficiency, ultimately resulting in minimal compute budget $C_{min}$:

\begin{equation}
\begin{split}
C_{min} = \frac{C}{1 + \frac{B}{B_{crit}(L)}}.
\end{split}
\end{equation}

Subsequently, we meticulously trained a series of MoE models, spanning from 10 M to 1B activation parameters, utilizing 100B tokens of pre-training data. By leveraging the isoFLOPs \citep{hoffmann2022training} curve, we determined the optimal number of active parameters and training data volume within a restricted compute budget, adjusted according to the actual training token batch size, through an exploration of these models across data sizes ranging from 10B to 100B tokens.

By fitting the formula $N_{opt} = N_c C_{min}^\alpha$ in Figure \ref{fig:IsoFLOP_curve-AP}, we obtain $N_c = 5.9 \times 10^{-3}$ and $\alpha = 0.5305$, indicating that the optimal number of activated parameters is approximately 58.1B. Inspired by \citet{dubey2024llama}, due to the smoothness of the quadratic curve around the optimal value, we ultimately select 52B as the number of activated parameters in our model.

\begin{figure}[!hbtp]
	\centering
	\includegraphics[width=0.88\columnwidth]{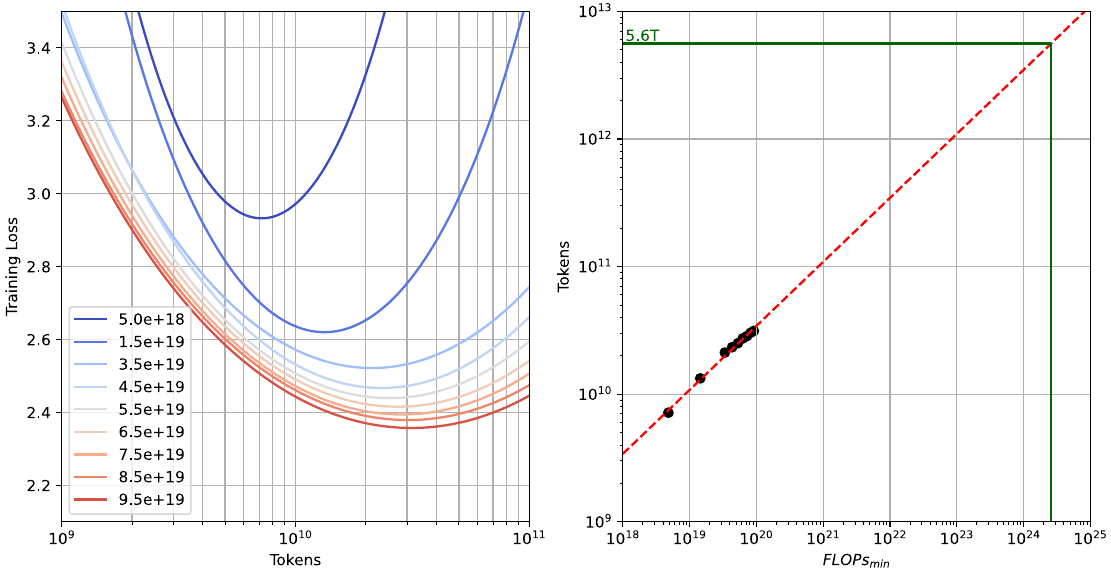}
	\caption{Employing the same fitting strategy as Figure \ref{fig:IsoFLOP_curve-AP}, we derive the scaling law of the optimal amount of training data under different minimum compute budgets.}
	\label{fig:IsoFLOP_curve-TK}
\end{figure}

Further, by fitting the formula $D_{opt} = D_c C_{min}^\beta$ in Figure \ref{fig:IsoFLOP_curve-TK}, we obtain $D_c = 3.2$ and $\beta = 0.50$, estimating the optimal number of trained tokens to be approximately 5.6T. Based on the same principle of smooth curves, and aiming to maximize the use of training data within the optimal cost-performance range to achieve the best possible model outcomes, we ultimately selected approximately 7T tokens for pre-training.
These analyses ensure that Hunyuan-Large achieves optimal performance at the best possible cost-efficiency, while also facilitating the development of a future series of MoE models.

\subsubsection{Learning Rate Scheduling}
An optimal learning rate schedule  is crucial for effective and stable training. Hunyuan-Large's learning rate schedule is delineated into three sequential phases: an initial warmup phase, succeeded by a prolonged phase of gradual decay, and culminating in a concise annealing phase.

The merit of the extended phase of gradual decay is its adeptness at balancing the exploration of the solution space with the convergence toward an optimal solution. By sustaining an elevated learning rate during the initial pre-training phase, the model is enabled to efficaciously navigate through diverse regions of the solution space, thereby averting premature convergence to suboptimal local minima. The incremental reduction in the learning rate as training progresses ensures a methodical convergence to a more optimal solution.

In the concluding 5\% of the pre-training tokens, we introduce a brief annealing phase, wherein the learning rate is reduced to one-tenth of its peak value. This approach facilitates the model in meticulously fine-tuning its parameters, thereby achieving a superior degree of generalization and, consequently, enhancing its overall performance. Furthermore, during this phase, we prioritize the use of the highest-quality dataset available, which plays a pivotal role in augmenting the model's performance in the annealing phase.

\subsubsection{Long-Context Pre-Training}
After the annealing phase, Hunyuan-Large is trained on longer sequences (up to 256K tokens) to enable its long-context capability. Specifically, the long-context pre-training phase contains two stages (i.e., gradually increases the token length as 32K$\rightarrow$256K). We adopt RoPE \citep{su2024roformer} for building position embeddings, and scale the RoPE base frequency to 1 billion during the 256K pre-training stage inspired by \cite{xiong2023effective}.

For the data, we solely rely on natural long-context data obtained from books and codes (comprising nearly $25\%$ of the corpus) and mix it with normal-length pre-training data (nearly $75\%$) to form our long-context pre-training corpus, sharing similar conclusions observed in \cite{gao2024train}. We also discover that it does not require too much training for LLM to acquire long-context capabilities. In each of the 32K and 256K stages, we employ a long-context pre-training corpus of approximately 10 billion tokens. The long-context pre-training at each stage can achieve satisfactory long-context abilities, while maintaining good LLM capabilities on tasks with normal lengths.

\section{Post-Training}
\label{sec:post-training}
Based on the pre-trained model of Hunyuan-Large, we further conduct a post-training stage that aims to enhance task-specific capabilities and align LLM to human preference. This stage contains a supervised fine-tuning (SFT) phase and a Reinforcement Learning from Human Feedback (RLHF) phase on elaborately selected datasets and outputs of current policy models \citep{bai2022training}. The following subsections contain (a) the data selection, preprocessing, and training process of SFT, (b) the techniques and training strategies of Direct Preference Optimization (DPO) in RLHF.

\subsection{Supervised Fine-Tuning}
The performance of SFT strongly depends on the quality of instruction data related to various types of LLM capabilities. In SFT, we concentrate on the detailed data collection and processing manners that ensure the effectiveness of Hunyuan-Large's post-training, along with the training settings of SFT.

\subsubsection{Overview of SFT Data}

The central goal of SFT is further enhancing its performance across multiple key capabilities based on the corresponding well-selected data. These capabilities primarily encompass mathematics, coding, logical reasoning, knowledge-based question answering, agent behavior, text generation, NLP comprehension, industrial applications, role-playing, long-text capabilities, etc. We recognize that improving these abilities not only enables the model to be more adept in practical applications, but also better satisfies users' diverse needs across multiple scenarios. Simultaneously, we place great emphasis on data security, striving to ensure that the model aligns with human values under most circumstances. The overall SFT data volume exceeds 1 million.

\subsubsection{Data Collection and Processing}

The key techniques of SFT data collection and processing mainly include instruction extraction, instruction generalization, instruction balancing, and data quality controlling.

\textbf{Instruction Extraction.}
To enhance the breadth and diversity of the instruction set, we develop an instruction extraction model specifically for domains such as mathematics, logical reasoning, and knowledge-based question answering, whose primary goal is to effectively extract data suitable for instruction tuning from publicly available data sources (e.g., web pages, encyclopedias, etc.). The extracted data includes both instructions and corresponding reference answers. We develop many specialized models as instruction extractors. With the help of these model, we successfully extract a large set of natural instructions from public data. These instructions play a crucial role as the seed to enhance the final model's generalization performance and diversity.

\textbf{Instruction Generalization.}
We propose an instruction generalization method to obtain more diverse and complex instructions in large quantities.
Specifically, we design and train an instruction generalization system capable of generalizing targeted instructions while gradually increasing their difficulty and complexity levels. The central recipe of this system lies in training the model by synthesizing numerous mappings between simple and complex instructions.
In addition, we construct a well-structured instruction taxonomy with its corresponding classification models, which aims to analyze and balance the distribution of various instruction types in SFT data. Armed with this instruction taxonomy, our instruction generalization system can supplement the original data on specific weak instructions of targeted types.

\textbf{Instruction Balancing.}
Through the instruction extraction and generalization processes, we accumulate more than $10$ million instructions. Instruction balance is essential for enhancing the model's performance across various scenarios. However, many generated instructions have very similar semantic meanings and the instruction type distribution is naturally unbalanced.
To enhance the instruction complexity while maintaining balanced instruction distributions, we attach labels for each instruction. These labels encompass multiple dimensions. By meticulously tagging these labels, we can more accurately understand and analyze the characteristics of our instruction sets. 
By ensuring adequate amounts and balanced distribution of different types of instructions during the SFT process, we can effectively alleviate overfitting or underfitting problems on specific instruction types, thereby improving the model's generalization capabilities and adaptability across diverse application scenarios.

\textbf{Data Quality Controlling}.
The quality of SFT data is the foundation of superior performance. We mainly conduct the following three methods to ensure the high quality of our SFT data.

\begin{itemize}[leftmargin=10pt]
    \item \textbf{Rule-based Filtering.}
    We discover some common issues such as data truncation errors, duplication, garbled characters, and format errors in SFT data. Consequently, we develop a set of rule-based data filtering strategies to prevent the above instruction extraction and generation models from producing undesirable outputs.
    \item \textbf{Model-based Filtering.}
    To automatically extract high-quality SFT data from a substantial volume of synthesized instruction data, we train a critique model \citep{mcaleese2024llm} based on a 70B dense model of our Hunyuan series. This model assigns a four-tier quality score to each instruction sample, assessing aspects such as the accuracy, relevance, completeness, usefulness, and clarity of the generated responses, and other possible data quality issues. 
    \item \textbf{Human-based Filtering.}
    Prior to model training, the SFT data filtered via rule-based and model-based methods further undergo human annotation, ensuring that answers adhere to the desired task-specific response patterns and avoid introducing additional low-quality issues.
\end{itemize}

\subsubsection{Training Details}
In SFT, we fine-tune the pre-trained model based on the high-quality data (more than 1 million) for a total of 3 epochs. The learning rate decays from 2e-5 to 2e-6. To mitigate overfitting during SFT, we utilize an attention dropout of 0.1 and a hidden dropout of 0.2. We find that, compared to the dense models, the MoE architecture of Hunyuan series could benefit more from incorporating suitable dropout rates.

\subsection{Reinforcement Learning from Human Feedback}

To align Hunyuan-Large with human preferences, we further train our SFT model using DPO \citep{rafailov2024direct}. We adopt a single-stage training strategy that integrates both offline and online training, which demonstrates superior controllability and overall performance. In this integrated approach, we utilize a pre-compiled preference dataset to enhance controllability, while simultaneously employing the current policy model to generate multiple responses for each prompt and our reward model to select the most and least preferred responses.

To enhance training stability, we incorporate an SFT loss term on the chosen response, similar to the approaches in \citep{dubey2024llama, adler2024nemotron}. This addition helps stabilize DPO training by preventing a decrease in the log probability of chosen responses.
Furthermore, we implement an exponential moving average strategy to mitigate reward hacking and reduce alignment tax \citep{ouyang2022training}, ensuring a more stable training process across a larger dataset.

\section{Model Evaluations} 
\label{sec:evaluation}

We conduct extensive evaluations of Hunyuan-Large to demonstrate its effectiveness. The following experiments concentrate on our pre-trained language model (in Sec. \ref{sec:pre_eval}) and post-trained language model (in Sec. \ref{sec:post_eval}) on various tasks in Chinese and English, including math and reasoning, code, reading comprehension, commonsense, long context, and aggregated task, etc., where Hunyuan-Large achieves excellent performance among tasks in both pre-training and post-training. 

\subsection{Evaluations on Pre-Trained Model}
\label{sec:pre_eval}
\begin{table}[!hbtp]
	\centering
        \caption{Performance of Hunyuan-Large's pre-trained model and its competitors.}
	\resizebox{0.99\linewidth}{!}{
	\begin{tabular}{lccccc}
\toprule
\textbf{Model}      & \textbf{LLama3.1-405B} & \textbf{LLama3.1-70B} & \textbf{Mixtral-8x22B} & \textbf{DeepSeek-V2} & \textbf{Hunyuan-Large} \\ \hline
Architecture        & Dense         & Dense        & MoE           & MoE         & MoE           \\
\# Activated Params & \textbf{405B}          & 70B          & 39B           & 21B         & 52B           \\
\# Total Params     & \textbf{405B}          & 70B          & 141B          & 236B        & 389B          \\
Context Length      & 128k          & 128k         & 64k           & 128k        & \textbf{256k}          \\ \hline
\multicolumn{6}{c}{English}                                                                      \\
MMLU                & 85.2          & 79.3         & 77.8          & 78.5        & \textbf{88.4} \\
MMLU-Pro            & \textbf{61.6} & 53.8         & 49.5          & -           & 60.2          \\
BBH                 & 85.9          & 81.6         & 78.9          & 78.9        & \textbf{86.3} \\
HellaSwag           & -             & -            & \textbf{88.7} & 87.8        & 86.8          \\
CommonsenseQA       & 85.8          & 84.1         & 82.4          & -           & \textbf{92.9} \\
WinoGrande          & 86.7          & 85.3         & 85.0          & 84.9        & \textbf{88.7} \\
PIQA                & -             & -            & 83.6          & 83.7        & \textbf{88.3} \\
NaturalQuestions    & -             & -            & 39.6          & 38.7        & \textbf{52.8} \\
DROP                & 84.8          & 79.6         & 80.4          & 80.1        & \textbf{88.9} \\
ARC-C               & \textbf{96.1} & 92.9         & 91.2          & 92.4        & 95.0          \\
TriviaQA            & -             & -            & 82.1          & 79.9        & \textbf{89.2} \\ \hline
\multicolumn{6}{c}{Chinese}                                                                      \\
CMMLU               & -             & -            & 60.0          & 84.0        & \textbf{90.2} \\
C-Eval              & -             & -            & 59.6          & 81.7        & \textbf{91.9} \\
C3                  & -             & -            & 71.4          & 77.4        & \textbf{82.3} \\ \hline
\multicolumn{6}{c}{Math}                                                                         \\
GSM8K               & 89.0          & 83.7         & 83.7          & 79.2        & \textbf{92.8} \\
MATH                & 53.8          & 41.4         & 42.5          & 43.6        & \textbf{69.8} \\
CMATH               & -             & -            & 72.3          & 78.7        & \textbf{91.3} \\ \hline
\multicolumn{6}{c}{Code}                                                                         \\
HumanEval           & 61.0          & 58.5            & 53.1          & 48.8        & \textbf{71.4} \\
MBPP                & \textbf{73.4}          & 68.6         & 64.2          & 66.6        & 72.6 \\ 
\bottomrule
\end{tabular}
	}
	\label{tab:main_pre_train}
\end{table}

In this section, we report the performance of Hunyuan-Large's pre-trained model on various types of widely-used benchmarks, verifying the power of the fundamental capability of our model.

\subsubsection{Benchmarks and Experimental Settings}

\textbf{Key Benchmarks.}
We evaluate Hunyuan-Large on a large number of widely-used benchmarks of various tasks, including commonsense understanding, machine comprehension, question answering, math and reasoning, coding, and aggregated tasks, in both English and Chinese.
Specifically, we choose MMLU \citep{hendrycks2021measuring}, MMLU-Pro \citep{wang2024mmlu-pro}, BBH \citep{suzgun2022challenging}, CMMLU \citep{li2023cmmlu}, and C-Eval \citep{huang2024ceval} for the aggregated evaluations.
HellaSwag \citep{zellers2019hellaswag}, CommonsenseQA \citep{talmor2019commonsenseqa}, and WinoGrande \citep{sakaguchi2021winogrande} are adopted to measure our model on commonsense understanding, while PIQA \citep{bisk2020piqa} is specially for physical related commonsense.
We also select DROP \citep{dua2019drop}, C3 \citep{sun2020investigating} and NaturalQuestions\citep{kwiatkowski2019natural} to evaluate LLMs on their basic capabilities on classical NLP tasks such as question answering and reading comprehension. 
ARC-C \citep{clark2018think}, TriviaQA \citep{joshi2017triviaqa} are added as QA tasks that require certain background related to science and updated world knowledge.
Finally, we evaluate LLMs on GSM8k \citep{cobbe2021training}, MATH \cite{hendrycks2021measuring}, and CMATH \citep{wei2023cmath} to measure the mathematics capability, and on HumanEval \citep{chen2021evaluating} and MBPP \citep{austin2021program} for coding, which are representative and essential LLM abilities.

\textbf{Evaluation Settings and Competitors.}
We follow the commonly used evaluation settings (e.g., evaluation metrics and the numbers of shots) of various types of benchmarks in experiments. Precisely, we adopt zero-shot for TriviaQA, PIQA, C3, HumanEval, 3-shot for BBH, MBPP, DROP, CMATH, 4-shot for GSM8K, MATH, 5-shot for MMLU, MMLU-Pro, C-Eval, CMMLU, WinoGrande, NaturalQuestions, 7-shot for CommonsenseQA, 10-shot for HellaSwag and 25-shot for ARC-C.
We compare Hunyuan-Large with the state-of-the-art Dense and MoE pre-trained models of comparable or larger (activated) parameter sizes. Specifically, these competitors include LLama3.1-70B \citep{dubey2024llama}, Mixtral-8x22B \citep{mistral2024mixtral8x22}, DeepSeek-V2 \citep{deepseek2024v2} and LLama3.1-405B. 
For fair comparisons, we report the best performance among the results that are publicly reported or those reproduced by ourselves for baselines.

\subsubsection{Model Performance of Pre-Training}

Table \ref{tab:main_pre_train} illustrates the performance of Hunyuan-Large and other competitive pre-trained models. In general, Hunyuan-Large achieves the best overall performance compared to both Dense and MoE based competitors having similar activated parameter sizes.
For aggregated benchmarks such as MMLU, Hunyuan-Large not only surpasses the performance of the LLama3.1-405B model but does so with a significantly lower count of activation parameters, achieving an impressive 3.2\% improvement. Hunyuan-Large also shows superior performance in commonsense understanding and reasoning, and classical NLP tasks such as QA and reading comprehension tasks (e.g., CommonsenseQA, PIQA, and TriviaQA). For the mathematics capability, Hunyuan-Large outperforms all baselines in math datasets of GSM8K and MATH, and also gains the best results on CMATH in Chinese. It also achieves the first-tier results in code datasets like HumanEval and MBPP. We also observe that Hunyuan-Large achieves the overall best performance in all Chinese tasks (e.g., CMMLU, C-Eval).
In-depth analyses throughout Hunyuan-Large's development process indicate that the impressive all-aspect improvement mainly derives from: (a) the high-quality pre-training data armed with synthesis techniques, functioning as the fundamental fuel of acquiring capabilities, (b) better model structure with recycle routing and expert-specific learning rate scaling on shared/specialized experts, and (c) the pre-training recipe inspired by various pioneer explorations on more effective and efficient MoE-based pre-training schedule, which enables a more intelligent and stable training. Furthermore, Hunyuan-Large is capable of handling longer sequences of up to 256K tokens attributed to our long-context pre-training. 

\subsection{Evaluations on Post-Trained Models}
\label{sec:post_eval}
\begin{table}[!hbtp]
	\centering
        \caption{Performance of our Hunyuan-Large-Instruct and its competitors. }
		\begin{tabular}{lccccc}
			\toprule
			\textbf{Model} & \makecell[c]{\textbf{LLama3.1}\\\textbf{405B Inst.}} & \makecell[c]{\textbf{LLama3.1}\\\textbf{70B Inst.}} & \makecell[c]{\textbf{Mixtral}\\ \textbf{8x22B Inst.}} & \makecell[c]{\textbf{DeepSeek}\\\textbf{V2.5 Chat}} & \makecell[c]{\textbf{Hunyuan-Large}\\\textbf{Inst.}} \\
        \midrule
MMLU                 & 87.3          & 83.6 & 77.8 & 80.4          & \textbf{89.9} \\
CMMLU                & -             & -    & 61.0 & -          & \textbf{90.4} \\
C-Eval               & -             & -    & 60.0 & -          & \textbf{88.6} \\
BBH                  & -             & -    & 78.4 & 84.3 & \textbf{89.5}          \\
ARC-C                & \textbf{96.9} & 94.8 & 90.0 & -          & 94.6          \\
GPQA\_diamond        & \textbf{51.1} & 46.7 & - & -          & 42.4          \\
MATH                 & 73.8          & 68.0 & 49.8 & 74.7          & \textbf{77.4} \\
HumanEval            & 89.0 & 80.5 & 75.0 & 89.0          & \textbf{90.0}         \\
\midrule
AlignBench           & 6.0             & 5.9    & 6.2  & 8.0             & \textbf{8.3}  \\
MT-Bench             & 9.1             & 8.8    & 8.1  & 9.0          & \textbf{9.4}  \\
IFEval strict-prompt & \textbf{86.0}             & 83.6    & 71.2 & -             & 85.0 \\
Arena-Hard & 69.3             & 55.7    & - & 76.2             & \textbf{81.8} \\
AlpacaEval-2.0 &  39.3 & 34.3 & 30.9 & 50.5 & \textbf{51.8}\\
        \bottomrule
	\end{tabular}
	\label{tab:main_post_train}
\end{table}
We present the results of the post-trained model of Hunyuan-Large, i.e., Hunyuan-Large-Instruct, on dozens of benchmarks to verify its effectiveness across different LLM capabilities.

\subsubsection{Benchmarks and Experimental Settings}

We directly adopt some benchmarks used in the evaluation of pre-training as the datasets to confirm the post-trained model's capability, which concentrate on the machine comprehension, question answering, commonsense reasoning, mathematics, coding, and aggregated tasks in both English and Chinese.
For fair comparisons, we follow the classical evaluation settings such as metrics and the numbers of shots for different benchmark datasets. As for baselines, we choose LLama3.1-405B-Instruct \citep{dubey2024llama}, LLama3.1-70B-Instruct, Mixtral-8x22B-Instruct \citep{mistral2024mixtral8x22}, and DeepSeek-V2.5-Chat \citep{deepseek2024v2}, which are powerful dense or MoE models with the similar (activated) parameter sizes. We follow the setting of the pre-training evaluation, reporting the best performance among the publicly reported scores and the results that we reproduced.

\subsubsection{Model Performance of Post-Training}
Table \ref{tab:main_post_train} shows the results of Hunyuan-Large-Instruct and its competitors on public benchmarks, from which we could observe that our Hunyuan-Large-Instruct achieves consistent improvements on most types of tasks compared to LLMs having similar activated parameters, indicating the effectiveness of our post-training. Delving into the model performance in different categories of benchmarks, we find that our instruct model achieves the best performance on MMLU and MATH dataset. Notably, on the MMLU dataset, our model demonstrates a significant improvement, outperforming the LLama3.1-405B model by 2.6\%. This enhancement is not just marginal but indicative of the Hunyuan-Large-Instruct's superior understanding and reasoning capabilities across a wide array of language understanding tasks. 
The model's prowess is further underscored in its performance on the MATH dataset, where it surpasses the LLama3.1-405B by a notable margin of 3.6\%. Remarkably, this leap in accuracy is achieved with only 52 billion activated parameters, underscoring the efficiency of our model.

In our pursuit to thoroughly evaluate the capabilities of Hunyuan-Large-Instruct, we further conduct comparisons on AlignBench \citep{liu2024alignbenchbenchmarkingchinesealignment}, MT-Bench \citep{zheng2023judgingllmasajudgemtbenchchatbot}, IFEval strict-prompt \citep{zhou2023instructionfollowingevaluationlargelanguage}, Arena-Hard \citep{li2024crowdsourceddatahighqualitybenchmarks}, and 
AlpacaEval-2.0 \citep{dubois2024lengthcontrolledalpacaevalsimpleway}, as Table \ref{tab:main_post_train} shows. (1). AlignBench is a benchmark designed to assess the alignment between model outputs and human intentions, particularly focusing on the model's ability to follow instructions accurately and generate responses that are in line with user expectations. (2). MT-Bench is an expert-level human preference benchmark for LLMs. (3). IFEval is a benchmark that specifically targets the model's performance in following strict prompts, thereby testing its precision and adherence to specific instructions within a given context. (4). Arena-Hard robustly differentiates model capabilities, aligns closely with human preferences in real-world scenarios, and is frequently updated with new prompts to prevent over-fitting and ensure ongoing relevance. (5). AlpacaEval-2.0 is also a commonly-used benchmark to automatically evaluate the LLMs' instruction following abilities. Hunyuan-Large-Instruct shows the best overall performance on these five benchmarks compared to all the strong baseline models. It is implied that the impressive performance of Hunyuan-Large-Instruct could mainly attribute to its powerful pre-trained model, the high-quality SFT and DPO data with the well-designed four-step data collection and processing that generates this data, and the superior SFT and DPO training strategies. 

\subsection{Long-Context Capability Evaluations}
To comprehensively assess the long-context performance of Hunyuan-Large-Instruct, we undertook a series of comprehensive assessments employing two widely recognized open-source benchmarks, i.e., RULER \citep{hsieh2024rulerwhatsrealcontext} and LV-Eval \citep{yuan2024lvevalbalancedlongcontextbenchmark}. In addition, we introduce a self-developed long-context benchmark, i.e., PenguinScrolls, for further comparisons. We select LLama3.1-70B-Instruct as a strong baseline due to its well-documented strength in processing extended contexts.

\subsubsection{Open-Source Long-Context Benchmarks and Evaluation}
\textbf{RULER}. RULER encompasses a diverse set of task categories, including retrieval, multi-hop reasoning, aggregation, and question answering. Each task spans varying context lengths, offering a flexible and comprehensive evaluation framework for assessing LLMs' long-context competencies. As depicted in Table \ref{tab:long-eval}, Hunyuan-Large-Instruct maintains consistently high performance across different lengths. Notably, in the 64K to 128K token range, Hunyuan-Large-Instruct significantly outperforms the baseline model, exhibiting minimal performance degradation with increasing length and demonstrating superior stability in handling extended text inputs.

\textbf{LV-Eval}. LV-Eval is a challenging long-context benchmark comprising $11$ distinct question-answering datasets, designed to test models across varying context lengths and challenging scenarios with confounding facts. To address the high false-negative rates caused by overly stringent original metrics, we employed LLM as an evaluator, providing a more accurate reflection of model performance. As illustrated in Table \ref{tab:long-eval}, Hunyuan-Large-Instruct consistently outperforms LLama3.1-70B-Instruct across all length intervals, underscoring its excellence in long-context processing. 

\begin{table}[!hbtp]
\centering
\caption{The performance of Hunyuan-Large-Instruct on RULER and LV-Eval.}
\resizebox{0.99\linewidth}{!}{
\begin{tabular}{c|cccc|ccc}
\hline
{ }                                            & \multicolumn{4}{c|}{{ \textbf{RULER}}}                                                                                                         & \multicolumn{3}{c}{{ \textbf{LV-Eval}}}                                                                \\ 
\multirow{-2}{*}{{\textbf{Model}}}            & {\textbf{0-8K}} & {\textbf{8K-32K}} & { \textbf{32K-64K}} & { \textbf{64K-128K}} & { \textbf{0-32K}} & { \textbf{32K-64K}} & { \textbf{64K-128K}} \\ \hline
{LLama3.1-70B-Instruct} & \textbf{95.89}                       & \textbf{95.39}                         & \textbf{94.72}                          & 86.48                                    & 75.73                        & 62.39                                   & 61.57                                    \\ 
Hunyuan-Large-Instruct                                             & 94.39                                & 94.94                                  & 93.02                                   & \textbf{89.53}                           & \textbf{81.92}                                 & \textbf{71.15}                          & \textbf{67.87}                           \\ \hline
\end{tabular}
}
\label{tab:long-eval}
\end{table}

\subsubsection{In-House Evaluation with PenguinScrolls}
To address the gaps in existing benchmarks, such as insufficient real-world content diversity and lack of multilingual and multi-turn dialogue data, we developed PenguinScrolls. 
This benchmark aims to guide the optimization of LLMs' long-text capabilities and align evaluation metrics with genuine user perceptions of LLMs' performance.

PenguinScrolls offers the following advantages: (1). Document diversity: Involves a wide range of natural long-form texts, including financial reports, legal documents, and academic papers, with contexts extending up to 128K tokens. (2). Fine-grained task types: Features multi-level tasks of varying difficulty, constructing a comprehensive task classification system rooted in long-context processing abilities. (3). Multi-turn dialogue data: Incorporates human-simulated questioning to create authentic long-context multi-turn dialogue scenarios. (4). Multilingual support: Provides data in both Chinese and English to meet the needs of multilingual applications.

\begin{table}[!th]
\caption{The performance of Hunyuan-Large-Instruct on PenguinScrolls.}
\resizebox{0.99\linewidth}{!}{
\begin{tabular}{cccccc}
\hline
\textbf{Model}         & \textbf{\begin{tabular}[c]{@{}c@{}}Information\\ Extraction\end{tabular}} & \textbf{\begin{tabular}[c]{@{}c@{}}Information\\ Localization\end{tabular}} & \textbf{\begin{tabular}[c]{@{}c@{}}Qualitative\\ Analysis\end{tabular}} & \textbf{\begin{tabular}[c]{@{}c@{}}Numerical\\ Reasoning\end{tabular}} & \multicolumn{1}{l}{\textbf{Overall}} \\ \hline
LLama3.1-70B-Instruct  & 82.51                                                                     & 69.70                                                                       & 75.77                                                                   & 49.52                                                                  & 69.37                       \\
Hunyuan-Large-Instruct & \textbf{91.14}                                                            & \textbf{89.56}                                                              & \textbf{92.78}                                                          & \textbf{67.46}                                                         & \textbf{85.23}              \\ \hline
\end{tabular}

}
\label{tab:long-eval-2}
\end{table}

PenguinScrolls is a high-quality dataset capable of effectively guiding the optimization of long-context processing capabilities. It encompasses four distinct tasks, i.e., information extraction, information localization, qualitative analysis and numerical reasoning. As shown in Table \ref{tab:long-eval-2}, Hunyuan-Large-Instruct demonstrates superior performance over LLama3.1-70B-Instruct across all these tasks. 
Internal user studies corroborate that the improvements on PenguinScrolls strongly correlate with enhancements in actual user experiences. We will release PenguinScrolls to advance long-context research and development in the future.

\section{Conclusions and Future Work}
\label{sec:conclusion}
This technical report presents Hunyuan-Large, the currently largest and best-performing Transformer-based MoE model, which has an unprecedented 389B total parameters and 52B activated parameters, being able to support up to 256K context length. Extensive evaluations demonstrate Hunyuan-Large's remarkable performance on dozens of benchmarks, reflecting its impressive LLM capabilities in language understanding, generation, reasoning, mathematics, coding, long context, and aggregated tasks. The favorable outcomes of our models are largely attributed to our high-quality training data armed with data synthesis, superior model structure, and sophisticated training recipes in both pre-training and post-training.
We have released the model weights of Hunyuan-Large to facilitate the community, looking forward to inspiring future research innovations and practical applications and achieving positive social impact. We also hope that the release of the largest and overall best-performing MoE-based Hunyuan-Large could spark more ripple of debate about more promising techniques of LLMs among the community, in turn to further improve our model from more practical aspects and contribute to the more helpful AGI in the future.

\section*{Contributors and Acknowledgements}
\label{sec:contributors}
Hunyuan-Large has seen significant participation and contributions from many teams at Tencent, for which we are deeply grateful. Here, we acknowledge the most central contributors involved in this project.

Xingwu Sun, Yanfeng Chen, Yiqing Huang, Ruobing Xie, Jiaqi Zhu, Kai Zhang, Shuaipeng Li, Zhen Yang, Jonny Han, Xiaobo Shu, Jiahao Bu, Zhongzhi Chen, Xuemeng Huang, Fengzong Lian, Saiyong Yang, Jianfeng Yan, Yuyuan Zeng, Xiaoqin Ren, Chao Yu, Lulu Wu, Yue Mao, Jun Xia, Tao Yang, Suncong Zheng, Kan Wu, Dian Jiao, Jinbao Xue, Xipeng Zhang, Decheng Wu, Kai Liu, 
Dengpeng Wu, Guanghui Xu, Shaohua Chen, Shuang Chen, Xiao Feng, 
Yigeng Hong, Junqiang Zheng, Chengcheng Xu, Zongwei Li, Xiong Kuang, Jianglu Hu, Yiqi Chen, Yuchi Deng, Guiyang Li, Ao Liu, Chenchen Zhang, Shihui Hu, Zilong Zhao, Zifan Wu, Yao Ding, Weichao Wang, Han Liu, Roberts Wang, Hao Fei, Peijie Yu, Ze Zhao, 
Xun Cao, Hai Wang, Fusheng Xiang, Mengyuan Huang, Zhiyuan Xiong, 
Bin Hu, Xuebin Hou, Lei Jiang, Jianqiang Ma, Jiajia Wu, Yaping Deng, Yi Shen, Qian Wang, Weijie Liu, 
Jie Liu, Meng Chen, Liang Dong, Weiwen Jia, Hu Chen, Feifei Liu, 
Rui Yuan, Huilin Xu, Zhenxiang Yan, Tengfei Cao, 
Zhichao Hu, Xinhua Feng, Dong Du, Tinghao Yu, Yangyu Tao, Feng Zhang, Jianchen Zhu, Chengzhong Xu, 
Xirui Li, Chong Zha, Wen Ouyang, Yinben Xia, Xiang Li, Zekun He, Rongpeng Chen, Jiawei Song, Ruibin Chen, Fan Jiang, Chongqing Zhao, Bo Wang, Hao Gong, Rong Gan, 
Winston Hu, Zhanhui Kang, Yong Yang, Yuhong Liu, Di Wang, and Jie Jiang. 

\newpage

\bibliography{main}

\begin{thebibliography}{62}
\providecommand{\natexlab}[1]{#1}
\providecommand{\url}[1]{\texttt{#1}}
\expandafter\ifx\csname urlstyle\endcsname\relax
  \providecommand{\doi}[1]{doi: #1}\else
  \providecommand{\doi}{doi: \begingroup \urlstyle{rm}\Url}\fi

\bibitem[Abdin et~al.(2024)Abdin, Jacobs, Awan, Aneja, Awadallah, Awadalla, Bach, Bahree, Bakhtiari, Behl, et~al.]{abdin2024phi3}
Abdin, M., Jacobs, S.~A., Awan, A.~A., Aneja, J., Awadallah, A., Awadalla, H., Bach, N., Bahree, A., Bakhtiari, A., Behl, H., et~al.
\newblock Phi-3 technical report: A highly capable language model locally on your phone.
\newblock \emph{arXiv preprint arXiv:2404.14219}, 2024.

\bibitem[Achiam et~al.(2023)Achiam, Adler, Agarwal, Ahmad, Akkaya, Aleman, Almeida, Altenschmidt, Altman, Anadkat, et~al.]{achiam2023gpt}
Achiam, J., Adler, S., Agarwal, S., Ahmad, L., Akkaya, I., Aleman, F.~L., Almeida, D., Altenschmidt, J., Altman, S., Anadkat, S., et~al.
\newblock {GPT-4} technical report.
\newblock \emph{arXiv preprint arXiv:2303.08774}, 2023.

\bibitem[Adler et~al.(2024)Adler, Agarwal, Aithal, Anh, Bhattacharya, Brundyn, Casper, Catanzaro, Clay, Cohen, et~al.]{adler2024nemotron}
Adler, B., Agarwal, N., Aithal, A., Anh, D.~H., Bhattacharya, P., Brundyn, A., Casper, J., Catanzaro, B., Clay, S., Cohen, J., et~al.
\newblock Nemotron-4 340b technical report.
\newblock \emph{arXiv preprint arXiv:2406.11704}, 2024.

\bibitem[Ainslie et~al.(2023)Ainslie, Lee-Thorp, de~Jong, Zemlyanskiy, Lebr{\'o}n, and Sanghai]{ainslie2023gqa}
Ainslie, J., Lee-Thorp, J., de~Jong, M., Zemlyanskiy, Y., Lebr{\'o}n, F., and Sanghai, S.
\newblock {GQA}: Training generalized multi-query transformer models from multi-head checkpoints.
\newblock In \emph{Proceedings of EMNLP}, 2023.

\bibitem[Austin et~al.(2021)Austin, Odena, Nye, Bosma, Michalewski, Dohan, Jiang, Cai, Terry, Le, et~al.]{austin2021program}
Austin, J., Odena, A., Nye, M., Bosma, M., Michalewski, H., Dohan, D., Jiang, E., Cai, C., Terry, M., Le, Q., et~al.
\newblock Program synthesis with large language models.
\newblock \emph{arXiv preprint arXiv:2108.07732}, 2021.

\bibitem[Bai et~al.(2022)Bai, Jones, Ndousse, Askell, Chen, DasSarma, Drain, Fort, Ganguli, Henighan, et~al.]{bai2022training}
Bai, Y., Jones, A., Ndousse, K., Askell, A., Chen, A., DasSarma, N., Drain, D., Fort, S., Ganguli, D., Henighan, T., et~al.
\newblock Training a helpful and harmless assistant with reinforcement learning from human feedback.
\newblock \emph{arXiv preprint arXiv:2204.05862}, 2022.

\bibitem[Bisk et~al.(2020)Bisk, Zellers, Gao, Choi, et~al.]{bisk2020piqa}
Bisk, Y., Zellers, R., Gao, J., Choi, Y., et~al.
\newblock {PIQA}: Reasoning about physical commonsense in natural language.
\newblock In \emph{Proceedings of AAAI}, 2020.

\bibitem[Brandon et~al.(2024)Brandon, Mishra, Nrusimha, Panda, and Kelly]{brandon2024reducing}
Brandon, W., Mishra, M., Nrusimha, A., Panda, R., and Kelly, J.~R.
\newblock Reducing transformer key-value cache size with cross-layer attention.
\newblock \emph{arXiv preprint arXiv:2405.12981}, 2024.

\bibitem[Chen et~al.(2021)Chen, Tworek, Jun, Yuan, Pinto, Kaplan, Edwards, Burda, Joseph, Brockman, et~al.]{chen2021evaluating}
Chen, M., Tworek, J., Jun, H., Yuan, Q., Pinto, H. P. D.~O., Kaplan, J., Edwards, H., Burda, Y., Joseph, N., Brockman, G., et~al.
\newblock Evaluating large language models trained on code.
\newblock \emph{arXiv preprint arXiv:2107.03374}, 2021.

\bibitem[Clark et~al.(2018)Clark, Cowhey, Etzioni, Khot, Sabharwal, Schoenick, and Tafjord]{clark2018think}
Clark, P., Cowhey, I., Etzioni, O., Khot, T., Sabharwal, A., Schoenick, C., and Tafjord, O.
\newblock {Think you have Solved Question Answering? Try ARC, the AI2 Reasoning Challenge}.
\newblock \emph{arXiv preprint arXiv:1803.05457}, 2018.

\bibitem[Cobbe et~al.(2021)Cobbe, Kosaraju, Bavarian, Chen, Jun, Kaiser, Plappert, Tworek, Hilton, Nakano, et~al.]{cobbe2021training}
Cobbe, K., Kosaraju, V., Bavarian, M., Chen, M., Jun, H., Kaiser, L., Plappert, M., Tworek, J., Hilton, J., Nakano, R., et~al.
\newblock Training verifiers to solve math word problems.
\newblock \emph{arXiv preprint arXiv:2110.14168}, 2021.

\bibitem[DeepSeek-AI(2024)]{deepseek2024v2}
DeepSeek-AI.
\newblock {DeepSeek-V2}: A strong, economical, and efficient mixture-of-experts language model.
\newblock \emph{arXiv preprint arXiv:2405.04434}, 2024.

\bibitem[Dua et~al.(2019)Dua, Wang, Dasigi, Stanovsky, Singh, and Gardner]{dua2019drop}
Dua, D., Wang, Y., Dasigi, P., Stanovsky, G., Singh, S., and Gardner, M.
\newblock {DROP}: A reading comprehension benchmark requiring discrete reasoning over paragraphs.
\newblock \emph{arXiv preprint arXiv:1903.00161}, 2019.

\bibitem[Dubey et~al.(2024)Dubey, Jauhri, Pandey, Kadian, Al-Dahle, Letman, Mathur, Schelten, Yang, Fan, et~al.]{dubey2024llama}
Dubey, A., Jauhri, A., Pandey, A., Kadian, A., Al-Dahle, A., Letman, A., Mathur, A., Schelten, A., Yang, A., Fan, A., et~al.
\newblock {The Llama 3 Herd of Models}.
\newblock \emph{arXiv preprint arXiv:2407.21783}, 2024.

\bibitem[Dubois et~al.(2024)Dubois, Galambosi, Liang, and Hashimoto]{dubois2024lengthcontrolledalpacaevalsimpleway}
Dubois, Y., Galambosi, B., Liang, P., and Hashimoto, T.~B.
\newblock Length-controlled alpacaeval: A simple way to debias automatic evaluators, 2024.
\newblock URL \url{https://arxiv.org/abs/2404.04475}.

\bibitem[Fedus et~al.(2022)Fedus, Zoph, and Shazeer]{fedus2022switch}
Fedus, W., Zoph, B., and Shazeer, N.
\newblock Switch transformers: Scaling to trillion parameter models with simple and efficient sparsity.
\newblock \emph{Journal of Machine Learning Research}, 2022.

\bibitem[Gao et~al.(2024)Gao, Wettig, Yen, and Chen]{gao2024train}
Gao, T., Wettig, A., Yen, H., and Chen, D.
\newblock How to train long-context language models (effectively).
\newblock \emph{arXiv preprint arXiv:2410.02660}, 2024.

\bibitem[Gemini et~al.(2023)Gemini, Anil, Borgeaud, Wu, Alayrac, Yu, Soricut, Schalkwyk, Dai, Hauth, et~al.]{Gemini2023gemini}
Gemini, T., Anil, R., Borgeaud, S., Wu, Y., Alayrac, J.-B., Yu, J., Soricut, R., Schalkwyk, J., Dai, A.~M., Hauth, A., et~al.
\newblock Gemini: a family of highly capable multimodal models.
\newblock \emph{arXiv preprint arXiv:2312.11805}, 2023.

\bibitem[Goyal et~al.(2017)Goyal, Doll{\'a}r, Girshick, Noordhuis, Wesolowski, Kyrola, Tulloch, Jia, and He]{goyal2017accurate}
Goyal, P., Doll{\'a}r, P., Girshick, R., Noordhuis, P., Wesolowski, L., Kyrola, A., Tulloch, A., Jia, Y., and He, K.
\newblock {Accurate, large minibatch SGD: Training Imagenet in 1 hour}.
\newblock \emph{arXiv preprint arXiv:1706.02677}, 2017.

\bibitem[Hendrycks et~al.(2021)Hendrycks, Burns, Kadavath, Arora, Basart, Tang, Song, and Steinhardt]{hendrycks2021measuring}
Hendrycks, D., Burns, C., Kadavath, S., Arora, A., Basart, S., Tang, E., Song, D., and Steinhardt, J.
\newblock Measuring mathematical problem solving with the math dataset.
\newblock \emph{arXiv preprint arXiv:2103.03874}, 2021.

\bibitem[Hoffmann et~al.(2022)Hoffmann, Borgeaud, Mensch, Buchatskaya, Cai, Rutherford, Casas, Hendricks, Welbl, Clark, et~al.]{hoffmann2022training}
Hoffmann, J., Borgeaud, S., Mensch, A., Buchatskaya, E., Cai, T., Rutherford, E., Casas, D. d.~L., Hendricks, L.~A., Welbl, J., Clark, A., et~al.
\newblock Training compute-optimal large language models.
\newblock \emph{arXiv preprint arXiv:2203.15556}, 2022.

\bibitem[Hsieh et~al.(2024)Hsieh, Sun, Kriman, Acharya, Rekesh, Jia, and Ginsburg]{hsieh2024rulerwhatsrealcontext}
Hsieh, C.-P., Sun, S., Kriman, S., Acharya, S., Rekesh, D., Jia, F., and Ginsburg, B.
\newblock {RULER}: What's the real context size of your long-context language models?
\newblock \emph{arXiv preprint arXiv:2404.06654}, 2024.

\bibitem[Huang et~al.(2024)Huang, Bai, Zhu, Zhang, Zhang, Su, Liu, Lv, Zhang, Fu, et~al.]{huang2024ceval}
Huang, Y., Bai, Y., Zhu, Z., Zhang, J., Zhang, J., Su, T., Liu, J., Lv, C., Zhang, Y., Fu, Y., et~al.
\newblock {C-Eval}: A multi-level multi-discipline chinese evaluation suite for foundation models.
\newblock In \emph{Proceedings of NeurIPS}, 2024.

\bibitem[Jamba et~al.(2024)Jamba, Lenz, Arazi, Bergman, Manevich, Peleg, Aviram, Almagor, Fridman, Padnos, et~al.]{team2024jamba}
Jamba, T., Lenz, B., Arazi, A., Bergman, A., Manevich, A., Peleg, B., Aviram, B., Almagor, C., Fridman, C., Padnos, D., et~al.
\newblock Jamba-1.5: Hybrid transformer-mamba models at scale.
\newblock \emph{arXiv preprint arXiv:2408.12570}, 2024.

\bibitem[Jiang et~al.(2024)Jiang, Sablayrolles, Roux, Mensch, Savary, Bamford, Chaplot, Casas, Hanna, Bressand, et~al.]{jiang2024mixtral}
Jiang, A.~Q., Sablayrolles, A., Roux, A., Mensch, A., Savary, B., Bamford, C., Chaplot, D.~S., Casas, D. d.~l., Hanna, E.~B., Bressand, F., et~al.
\newblock Mixtral of experts.
\newblock \emph{arXiv preprint arXiv:2401.04088}, 2024.

\bibitem[Joshi et~al.(2017)Joshi, Choi, Weld, and Zettlemoyer]{joshi2017triviaqa}
Joshi, M., Choi, E., Weld, D.~S., and Zettlemoyer, L.
\newblock {TriviaQA}: A large scale distantly supervised challenge dataset for reading comprehension.
\newblock \emph{arXiv preprint arXiv:1705.03551}, 2017.

\bibitem[Kaplan et~al.(2020)Kaplan, McCandlish, Henighan, Brown, Chess, Child, Gray, Radford, Wu, and Amodei]{kaplan2020scaling}
Kaplan, J., McCandlish, S., Henighan, T., Brown, T.~B., Chess, B., Child, R., Gray, S., Radford, A., Wu, J., and Amodei, D.
\newblock Scaling laws for neural language models.
\newblock \emph{arXiv preprint arXiv:2001.08361}, 2020.

\bibitem[Krizhevsky(2014)]{krizhevsky2014one}
Krizhevsky, A.
\newblock One weird trick for parallelizing convolutional neural networks.
\newblock \emph{arXiv preprint arXiv:1404.5997}, 2014.

\bibitem[Kwiatkowski et~al.(2019)Kwiatkowski, Palomaki, Redfield, Collins, Parikh, Alberti, Epstein, Polosukhin, Devlin, Lee, et~al.]{kwiatkowski2019natural}
Kwiatkowski, T., Palomaki, J., Redfield, O., Collins, M., Parikh, A., Alberti, C., Epstein, D., Polosukhin, I., Devlin, J., Lee, K., et~al.
\newblock Natural questions: a benchmark for question answering research.
\newblock \emph{TACL}, 2019.

\bibitem[Lepikhin et~al.(2020)Lepikhin, Lee, Xu, Chen, Firat, Huang, Krikun, Shazeer, and Chen]{lepikhin2020gshard}
Lepikhin, D., Lee, H., Xu, Y., Chen, D., Firat, O., Huang, Y., Krikun, M., Shazeer, N., and Chen, Z.
\newblock {GShard}: Scaling giant models with conditional computation and automatic sharding.
\newblock \emph{arXiv preprint arXiv:2006.16668}, 2020.

\bibitem[Li et~al.(2023)Li, Zhang, Koto, Yang, Zhao, Gong, Duan, and Baldwin]{li2023cmmlu}
Li, H., Zhang, Y., Koto, F., Yang, Y., Zhao, H., Gong, Y., Duan, N., and Baldwin, T.
\newblock {CMMLU}: Measuring massive multitask language understanding in chinese.
\newblock \emph{arXiv preprint arXiv:2306.09212}, 2023.

\bibitem[Li et~al.(2024{\natexlab{a}})Li, Zhao, Zhang, Sun, Wu, Jiao, Wang, Liu, Fang, Xue, et~al.]{li2024surge}
Li, S., Zhao, P., Zhang, H., Sun, X., Wu, H., Jiao, D., Wang, W., Liu, C., Fang, Z., Xue, J., et~al.
\newblock Surge phenomenon in optimal learning rate and batch size scaling.
\newblock \emph{arXiv preprint arXiv:2405.14578}, 2024{\natexlab{a}}.

\bibitem[Li et~al.(2024{\natexlab{b}})Li, Chiang, Frick, Dunlap, Wu, Zhu, Gonzalez, and Stoica]{li2024crowdsourceddatahighqualitybenchmarks}
Li, T., Chiang, W.-L., Frick, E., Dunlap, L., Wu, T., Zhu, B., Gonzalez, J.~E., and Stoica, I.
\newblock From crowdsourced data to high-quality benchmarks: Arena-hard and benchbuilder pipeline, 2024{\natexlab{b}}.
\newblock URL \url{https://arxiv.org/abs/2406.11939}.

\bibitem[Liu et~al.(2024)Liu, Wei, Liu, Si, Zhang, Rao, Zheng, Peng, Yang, Zhou, et~al.]{liu2024best}
Liu, R., Wei, J., Liu, F., Si, C., Zhang, Y., Rao, J., Zheng, S., Peng, D., Yang, D., Zhou, D., et~al.
\newblock Best practices and lessons learned on synthetic data.
\newblock In \emph{Proceedings of COLM}, 2024.

\bibitem[Liu et~al.(2023)Liu, Lei, Wang, Huang, Feng, Wen, Cheng, Ke, Xu, Tam, et~al.]{liu2024alignbenchbenchmarkingchinesealignment}
Liu, X., Lei, X., Wang, S., Huang, Y., Feng, Z., Wen, B., Cheng, J., Ke, P., Xu, Y., Tam, W.~L., et~al.
\newblock {AlignBench}: Benchmarking chinese alignment of large language models.
\newblock \emph{arXiv preprint arXiv:2311.18743}, 2023.

\bibitem[Loshchilov \& Hutter(2019)Loshchilov and Hutter]{loshchilov2019decoupled}
Loshchilov, I. and Hutter, F.
\newblock Decoupled weight decay regularization.
\newblock In \emph{Proceedings of ICLR}, 2019.

\bibitem[McAleese et~al.(2024)McAleese, Pokorny, Uribe, Nitishinskaya, Trebacz, and Leike]{mcaleese2024llm}
McAleese, N., Pokorny, R.~M., Uribe, J. F.~C., Nitishinskaya, E., Trebacz, M., and Leike, J.
\newblock {LLM Critics Help Catch LLM Bugs}.
\newblock \emph{arXiv preprint arXiv:2407.00215}, 2024.

\bibitem[Mistral(2024)]{mistral2024mixtral8x22}
Mistral.
\newblock Cheaper, better, faster, stronger. continuing to push the frontier of {AI} and making it accessible to all.
\newblock 2024.
\newblock URL \url{https://mistral.ai/news/mixtral-8x22b}.

\bibitem[OpenAI(2022)]{OpenAI2022ChatGPT}
OpenAI.
\newblock {Introducing ChatGPT}.
\newblock 2022.
\newblock URL \url{https://openai.com/index/chatgpt/}.

\bibitem[OpenAI(2023)]{OpenAI2022Tiktoken}
OpenAI.
\newblock {Tiktoken}.
\newblock 2023.
\newblock URL \url{https://github.com/openai/tiktoken}.

\bibitem[OpenAI(2024)]{OpenAI2024GPT4o}
OpenAI.
\newblock {Hello GPT-4o}.
\newblock 2024.
\newblock URL \url{https://openai.com/index/hello-gpt-4o/}.

\bibitem[Ouyang et~al.(2022)Ouyang, Wu, Jiang, Almeida, Wainwright, Mishkin, Zhang, Agarwal, Slama, Ray, et~al.]{ouyang2022training}
Ouyang, L., Wu, J., Jiang, X., Almeida, D., Wainwright, C., Mishkin, P., Zhang, C., Agarwal, S., Slama, K., Ray, A., et~al.
\newblock Training language models to follow instructions with human feedback.
\newblock In \emph{Proceedings of NeurIPS}, 2022.

\bibitem[Qwen(2024{\natexlab{a}})]{QWen2024qwen2-5}
Qwen.
\newblock {Qwen2.5}.
\newblock 2024{\natexlab{a}}.
\newblock URL \url{https://github.com/QwenLM/Qwen2.5}.

\bibitem[Qwen(2024{\natexlab{b}})]{qwen_moe}
Qwen.
\newblock {Qwen1.5-MoE: Matching 7B Model Performance with 1/3 Activated Parameters}, 2024{\natexlab{b}}.
\newblock URL \url{https://qwenlm.github.io/blog/qwen-moe/}.

\bibitem[Rafailov et~al.(2024)Rafailov, Sharma, Mitchell, Manning, Ermon, and Finn]{rafailov2024direct}
Rafailov, R., Sharma, A., Mitchell, E., Manning, C.~D., Ermon, S., and Finn, C.
\newblock Direct preference optimization: Your language model is secretly a reward model.
\newblock In \emph{Proceedings of NeurIPS}, 2024.

\bibitem[Sakaguchi et~al.(2021)Sakaguchi, Bras, Bhagavatula, and Choi]{sakaguchi2021winogrande}
Sakaguchi, K., Bras, R.~L., Bhagavatula, C., and Choi, Y.
\newblock {WinoGrande}: An adversarial winograd schema challenge at scale.
\newblock \emph{Communications of the ACM}, 2021.

\bibitem[Shazeer(2020)]{shazeer2020glu}
Shazeer, N.
\newblock {GLU} variants improve transformer.
\newblock \emph{arXiv preprint arXiv:2002.05202}, 2020.

\bibitem[Su et~al.(2024)Su, Ahmed, Lu, Pan, Bo, and Liu]{su2024roformer}
Su, J., Ahmed, M., Lu, Y., Pan, S., Bo, W., and Liu, Y.
\newblock Roformer: Enhanced transformer with rotary position embedding.
\newblock \emph{Neurocomputing}, 2024.

\bibitem[Sun et~al.(2020)Sun, Yu, Yu, and Cardie]{sun2020investigating}
Sun, K., Yu, D., Yu, D., and Cardie, C.
\newblock Investigating prior knowledge for challenging chinese machine reading comprehension.
\newblock \emph{TACL}, 2020.

\bibitem[Suzgun et~al.(2022)Suzgun, Scales, Sch{\"a}rli, Gehrmann, Tay, Chung, Chowdhery, Le, Chi, Zhou, et~al.]{suzgun2022challenging}
Suzgun, M., Scales, N., Sch{\"a}rli, N., Gehrmann, S., Tay, Y., Chung, H.~W., Chowdhery, A., Le, Q.~V., Chi, E.~H., Zhou, D., et~al.
\newblock Challenging big-bench tasks and whether chain-of-thought can solve them.
\newblock \emph{arXiv preprint arXiv:2210.09261}, 2022.

\bibitem[Talmor et~al.(2019)Talmor, Herzig, Lourie, and Berant]{talmor2019commonsenseqa}
Talmor, A., Herzig, J., Lourie, N., and Berant, J.
\newblock {CommonsenseQA}: A question answering challenge targeting commonsense knowledge.
\newblock In \emph{Proceedings of NAACL}, 2019.

\bibitem[Touvron et~al.(2023)Touvron, Lavril, Izacard, Martinet, Lachaux, Lacroix, Rozi{\`e}re, Goyal, Hambro, Azhar, et~al.]{touvron2023llama}
Touvron, H., Lavril, T., Izacard, G., Martinet, X., Lachaux, M.-A., Lacroix, T., Rozi{\`e}re, B., Goyal, N., Hambro, E., Azhar, F., et~al.
\newblock {LLaMA: Open and Efficient Foundation Language Models}.
\newblock \emph{arXiv preprint arXiv:2302.13971}, 2023.

\bibitem[Vaswani et~al.(2017)Vaswani, Shazeer, Parmar, Uszkoreit, Jones, Gomez, Kaiser, and Polosukhin]{vaswani2017attention}
Vaswani, A., Shazeer, N., Parmar, N., Uszkoreit, J., Jones, L., Gomez, A.~N., Kaiser, L., and Polosukhin, I.
\newblock Attention is all you need.
\newblock In \emph{Proceedings of NIPS}, 2017.

\bibitem[Wang et~al.(2024{\natexlab{a}})Wang, Sun, Xie, Li, Zhu, Yang, Zhao, Han, Kang, Wang, et~al.]{wang2024hmoe}
Wang, A., Sun, X., Xie, R., Li, S., Zhu, J., Yang, Z., Zhao, P., Han, J., Kang, Z., Wang, D., et~al.
\newblock {HMoE}: Heterogeneous mixture of experts for language modeling.
\newblock \emph{arXiv preprint arXiv:2408.10681}, 2024{\natexlab{a}}.

\bibitem[Wang et~al.(2024{\natexlab{b}})Wang, Ma, Zhang, Ni, Chandra, Guo, Ren, Arulraj, He, Jiang, et~al.]{wang2024mmlu-pro}
Wang, Y., Ma, X., Zhang, G., Ni, Y., Chandra, A., Guo, S., Ren, W., Arulraj, A., He, X., Jiang, Z., et~al.
\newblock {MMLU-Pro}: A more robust and challenging multi-task language understanding benchmark.
\newblock In \emph{Proceedings of NeurIPS}, 2024{\natexlab{b}}.

\bibitem[Wei et~al.(2023)Wei, Luan, Liu, Dong, and Wang]{wei2023cmath}
Wei, T., Luan, J., Liu, W., Dong, S., and Wang, B.
\newblock {CMATH}: Can your language model pass chinese elementary school math test?
\newblock \emph{arXiv preprint arXiv:2306.16636}, 2023.

\bibitem[Xiong et~al.(2023)Xiong, Liu, Molybog, Zhang, Bhargava, Hou, Martin, Rungta, Sankararaman, Oguz, et~al.]{xiong2023effective}
Xiong, W., Liu, J., Molybog, I., Zhang, H., Bhargava, P., Hou, R., Martin, L., Rungta, R., Sankararaman, K.~A., Oguz, B., et~al.
\newblock Effective long-context scaling of foundation models.
\newblock \emph{arXiv preprint arXiv:2309.16039}, 2023.

\bibitem[Yang et~al.(2024)Yang, Yang, Hui, Zheng, Yu, Zhou, Li, Li, Liu, Huang, et~al.]{yang2024qwen2}
Yang, A., Yang, B., Hui, B., Zheng, B., Yu, B., Zhou, C., Li, C., Li, C., Liu, D., Huang, F., et~al.
\newblock {QWen2 Technical Report}.
\newblock \emph{arXiv preprint arXiv:2407.10671}, 2024.

\bibitem[Yuan et~al.(2024)Yuan, Ning, Zhou, Yang, Li, Zhuang, Tan, Yao, Lin, Li, et~al.]{yuan2024lvevalbalancedlongcontextbenchmark}
Yuan, T., Ning, X., Zhou, D., Yang, Z., Li, S., Zhuang, M., Tan, Z., Yao, Z., Lin, D., Li, B., et~al.
\newblock {LV-Eval}: A balanced long-context benchmark with 5 length levels up to 256k.
\newblock \emph{arXiv preprint arXiv:2402.05136}, 2024.

\bibitem[Zellers et~al.(2019)Zellers, Holtzman, Bisk, Farhadi, and Choi]{zellers2019hellaswag}
Zellers, R., Holtzman, A., Bisk, Y., Farhadi, A., and Choi, Y.
\newblock {HellaSwag}: Can a machine really finish your sentence?
\newblock In \emph{Proceedings of ACL}, 2019.

\bibitem[Zheng et~al.(2023)Zheng, Chiang, Sheng, Zhuang, Wu, Zhuang, Lin, Li, Li, Xing, et~al.]{zheng2023judgingllmasajudgemtbenchchatbot}
Zheng, L., Chiang, W.-L., Sheng, Y., Zhuang, S., Wu, Z., Zhuang, Y., Lin, Z., Li, Z., Li, D., Xing, E., et~al.
\newblock {Judging LLM-as-a-Judge with MT-Bench and Chatbot Arena}.
\newblock In \emph{Proceedings of NeurIPS}, 2023.

\bibitem[Zhou et~al.(2023)Zhou, Lu, Mishra, Brahma, Basu, Luan, Zhou, and Hou]{zhou2023instructionfollowingevaluationlargelanguage}
Zhou, J., Lu, T., Mishra, S., Brahma, S., Basu, S., Luan, Y., Zhou, D., and Hou, L.
\newblock Instruction-following evaluation for large language models.
\newblock \emph{arXiv preprint arXiv:2311.07911}, 2023.

\end{thebibliography}
\bibliographystyle{main}
\end{document}